\documentclass[letterpaper]{article} 
\PassOptionsToPackage{hidelinks}{hyperref}
\usepackage{aaai2026}  

\usepackage{times}  
\usepackage{helvet}  
\usepackage{courier}  
\usepackage[hyphens]{url}  
\usepackage{url}

\usepackage{graphicx} 
\usepackage{natbib}  
\usepackage{caption} 
\usepackage{algorithm}
\usepackage{algorithmic}
\usepackage{mdframed}
\usepackage{tcolorbox}
\tcbuselibrary{breakable, skins}
\usepackage{graphicx}
\usepackage{subcaption}
\usepackage{xcolor}
\usepackage{amsmath}
\usepackage{booktabs}
\usepackage{newfloat}
\usepackage{listings}
\DeclareCaptionStyle{ruled}{labelfont=normalfont,labelsep=colon,strut=off} 
\floatstyle{ruled}
\newfloat{listing}{tb}{lst}{}
\floatname{listing}{Listing}
\usepackage{xcolor}
\usepackage{hyperref}

\title{When Cow Urine Cures Constipation on YouTube: Limits of LLMs in Detecting Culture-specific Health Misinformation}

\author{
 Anamta Khan\equalcontrib \textsuperscript{\rm 1}, Ratna Kandala\equalcontrib\textsuperscript{\rm 2} , Deepti\textsuperscript{\rm 3} , Sheza Munir\textsuperscript{\rm 1}, Joyojeet Pal\textsuperscript{\rm 1} 
}
\affiliations{
    \textsuperscript{\rm 1} University of Michigan, \textsuperscript{\rm 2}University of Kansas, \textsuperscript{\rm 3}IIT Jodhpur
    
}

\usepackage{bibentry}
\usepackage[hidelinks]{hyperref}

\begin{document}

\maketitle

\begin{abstract}
Social media platforms have become primary channels for health information in the Global South. Using \textit{gomutra} (cow urine) discourse on YouTube in India as a case study, we present a post-facto Large Language Model (LLM)-assisted discourse analysis of 30 multilingual transcripts showing that promotional content blends sacred traditional language with pseudo-scientific claims in ways that sophisticated debunking content itself mirrors, creating a rhetorical register that LLMs, trained predominantly on Western corpora, are systematically ill-equipped to analyse. Varying prompt tone across three LLMs (GPT-4o, Gemini 2.5 Pro, DeepSeek-V3.1), we find that culturally embedded health misinformation does not look like ordinary misinformation, and this cultural obfuscation extends to gendered rhetoric and prompt design, compounding analytical unreliability. Our findings argue that cultural competency in LLM-assisted discourse analysis cannot be retrofitted through prompt engineering alone.
\end{abstract}

\section{Introduction}

Nearly sixty-four percent of the global population actively uses social media (5.2 billion), with 491 million in India alone \cite{Kemp2025DigitalGlobal}. Among these, YouTube use in India alone has increased by 29 million (+6.3\%) between January 2024 and 2025 \cite{Kemp2025DigitalIndia},  making it one of the primary channels for health information, and increasingly health misinformation \cite{li2022youtube}, posing a growing threat to public health \cite{vosoughi2018spread, RODRIGUES2024100846}, especially when it is embedded in cultural tradition \cite{Madathil2015YouTubeHealth} and endorsed by figures of authority \cite{reuters2024ramdevmisled, scio2020fighting, reuters2020trumpdisinfectant}. 

LLMs have shown genuine promise as tools for detecting and moderating health misinformation \cite{chen2024combating, Alarabid2025LLMMisinformationDetection}, with 17\% of the U.S. adults now consulting health chatbots monthly \cite{kff2024aihealth}. Recent work has extended this to accessing health content quality on YouTube directly \cite{khalil2025evaluating}. Yet this promise is unevenly distributed: LLM accuracy on health claim verification is systematically higher in English and European languages than in non-European ones, with performance varying substantially by topic and cultural context \cite{garg2025context}. Because LLMs generate text through next-token prediction over training corpora, health misinformation embedded upstream can be surfaced and amplified without any corrective mechanism. This risk is compounded by a systematic evaluation gap: 95\% of 519 LLM health evaluations conducted between 2022 and 2024 were in English, rarely assessing bias or fairness across other languages \cite{bedi2025testing}, and where multilingual performance has been tested, correctness and consistency dropped markedly outside English \cite{jin2024better,  garg2025context}.

This limitation is especially acute on YouTube, where the informal, conversational format allows speakers to draw on tone, facial expression, and personal testimony to build audience trust in ways text-based platforms do not. These rhetorical resources become particularly powerful when health claims are framed through shared cultural and religious identity, especially in India, creating a category of misinformation that is not straightforwardly false but persuades through cultural authority.

\begin{figure*}[!t]
    \centering
        \centering
        \includegraphics[width=\linewidth]{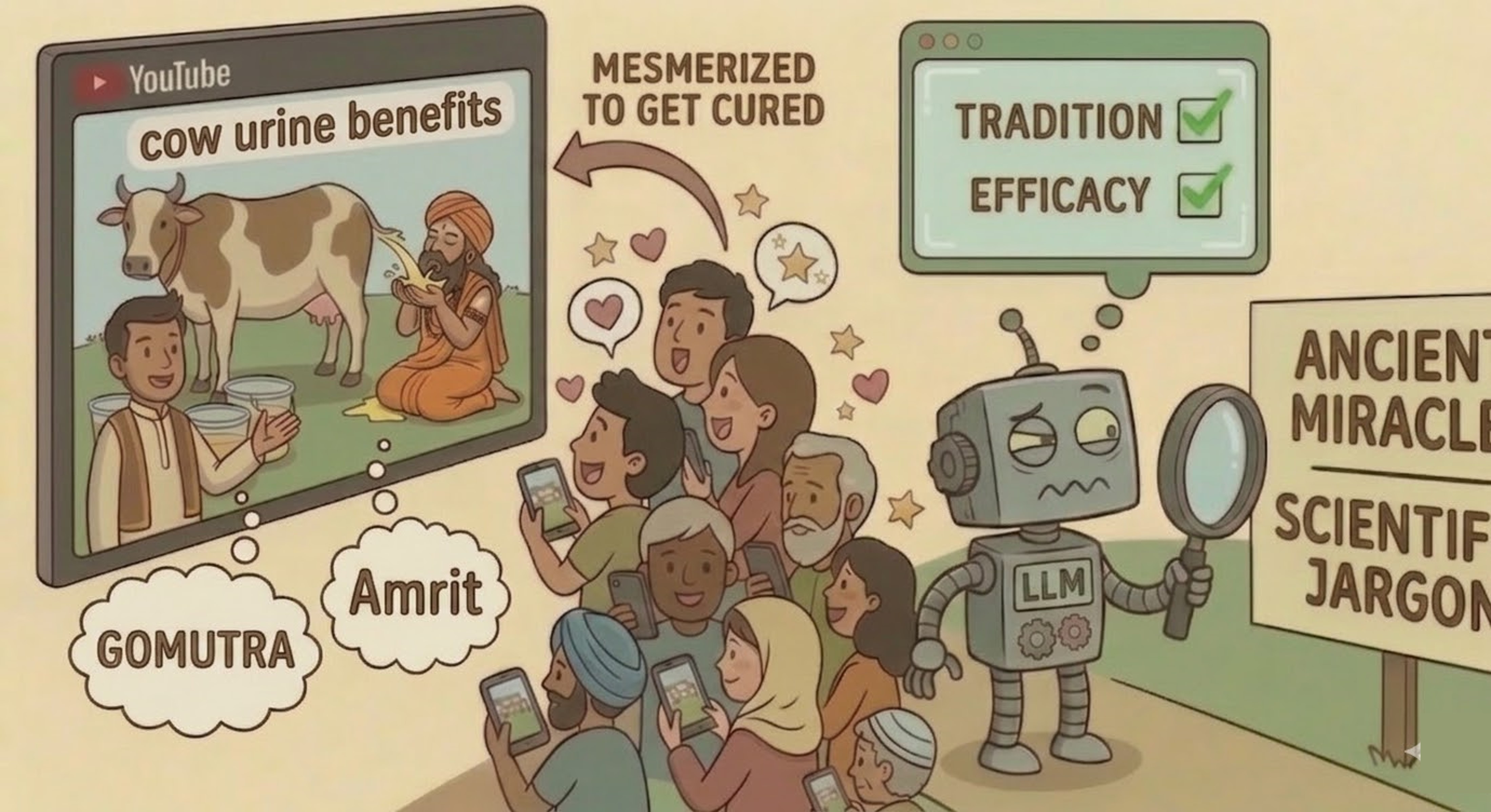}

    \caption{ 
An infographic visualizing the 'Digital Proliferation of Traditional Health Beliefs' through cow urine (\textit{gomutra}) remedies. The diagram contrasts the viral social media uptake against the challenge for Large Language Models (LLMs) to critically parse blended narratives that conflate ancient claims with pseudo-scientific terminology.
    }
    \label{fig:schematic_diagram}
\end{figure*}

One such claim centers on \textit{gomutra} (cow urine), a substance that holds deep religious significance in Hinduism, where cows are considered to be sacred, and their byproducts - including cow dung, urine, and milk are traditionally regarded as purifying and antiseptic agents \cite{Notermans2019Prayers, Essar2021CowDungCOVID}. The widespread adoption of cow urine as a remedy has been fueled by misinformation endorsed by politicians, authorities, and government bodies through media interviews, speeches, and official documents. A notable example is Indian Minister of State for Health Ashwini Choubey's 2019 public endorsement of cow urine as a basis for developing cancer treatments \cite{GulfNews2019CowUrine, indiatoday2019cowurine, Daria2021CowDungCOVID, indiatoday2022choubeydesai}. These endorsements have lent institutional legitimacy to health claims that lack scientific support, making them substantially harder to counter - whether through debunking, regulatory intervention, or platform moderation (Fig \ref{fig:schematic_diagram}). 

Yet despite this scale and urgency, discourse on \textit{gomutra}-related health claims remains significantly underexplored computationally. No annotated datasets exist, and no systematic linguistic analyses have been conducted. Existing work on Indian health misinformation has focused largely on COVID-19 claims on WhatsApp \cite{Essar2021CowDungCOVID} or on general misinformation diffusion patterns, without analyzing rhetorical structure. This exploratory study is a first attempt to address that gap directly.

We investigate how promoting and debunking content on \textit{gomutra} health claims differ in their linguistic and rhetorical strategies- specifically, how traditional metaphors and culturally resonant terminology function differently from scientific framing - using LLMs as analytical instruments to examine these patterns across a corpus of 30 YouTube transcripts. We employ three LLMs (GPT-4o, Gemini 2.5 Pro, and DeepSeek-V3.1) by systematically varying prompt tone (formal vs. informal/conversational) and vocabulary (traditional terms such as "Sanjivani" vs. scientific terminology) to evaluate how sensitive these models' outputs are to cultural framing. In doing so, we contribute to the broader literature on computational discourse analysis and cultural NLP by reflecting critically on using LLMs as viable tools in culturally specific health misinformation contexts - attending to both their analytical affordances and their blind spots.

In this regard, we investigate the following research questions:

\begin{enumerate}
    \item[\textbf{RQ1}] How do linguistic and rhetorical strategies (i.e., deliberate choices in language use, framing, and discourse structure that serve communicative goals) differ between promotional and debunking \textit{gomutra} content?
    
    \item[\textbf{RQ2}] How does prompt design, specifically tone, affect LLM sensitivity to linguistic markers of persuasion (i.e., lexical and rhetorical features associated with attitude change and influence) in \textit{gomutra} content?
    
    \item[\textbf{RQ3}] What do LLM sensitivities to cultural framing, gender, and prompt design reveal about their reliability as instruments in culturally embedded health misinformation contexts?
\end{enumerate}


Our findings reveal that promotional and debunking narratives deploy fundamentally different rhetorical strategies, and that LLMs' outputs are shaped by cultural framing, prompt design, and gendered rhetoric in ways that have direct implications for their use as discourse analytical instruments. This work sits at the intersection of NLP, health communication, and misinformation studies, and contributes from multiple angles: (a) empirically, a first attempt at an annotated video-transcript corpus of \textit{gomutra} health claims; (b) analytically, a linguistic account of persuasion strategies in this domain; and (c) ethically, a set of methodological cautions for researchers using Western LLMs \cite{adilazuarda2024towards} as analytical instruments in ways that risk systematic misrepresentation of non-Western culturally embedded discourse, with direct implications for fairness in any downstream use of these tools.


\section{Previous Work}

Recent research on social media misinformation has increasingly focused on high-stakes domains such as health \cite{kong2021tiktok}, where false claims pose a direct threat to societal wellbeing \cite{2025MisinformationPublicHealth}. Studies have also characterized the prevalence, spread, and psychological impact of health misinformation across these digital platforms \cite{VanDerLinden2022Misinformation,VanDerLinden2025HealthMisinformation}. Existing literature reveals that health misinformation employs distinct persuasive strategies, such as fabricating narratives, politicizing public health issues, and misappropriating scientific evidence, to legitimize false claims \cite{Peng2023PersuasiveHealthMisinformation}.

While LLMs excel in standard evaluation frameworks, they face significant challenges when confronted with misinformation that is not straightforwardly false, but instead persuades through culturally resonant rhetoric, traditional authority, and shared identity. To deconstruct how such rhetoric operates, researchers increasingly rely on linguistic analysis. For example, topical and linguistic analyses of COVID-19 misinformation on YouTube have identified dominant themes centered around conspiracy theories and political dissemination \cite{Thakur2024YouTubeMisinformation}. As generative AI reshapes this landscape, recent evaluations of Chinese datasets have mapped specific linguistic features of misinformation, including distinct patterns in sentiment and cognitive framing, while simultaneously identifying the detection limits of standard LLMs \cite{Ma2025AIMisinformationLinguistic}. Despite these insights, the majority of LLM development and evaluation remains focused on Western contexts. Consequently, these models suffer from a pervasive lack of cultural awareness \cite{Pawar2025CulturalAwarenessSurvey}, exhibiting inherent biases that severely limit their ability to parse culturally embedded rhetoric and implicit societal cues \cite{Liu2025CulturalBiasLLM}.

This cultural limitation is particularly pronounced in the Indian context, where empirical evaluations demonstrate that LLMs consistently fail to accurately parse culture-specific traditions and regional dialects \cite{Chhikara2025PrismCulture}. Furthermore, these models frequently exhibit stereotypical biases and struggle to comprehend complex subcultures \cite{Khandelwal2024IndianBhED}. This specific cultural complexity in detecting misinformation is clearly visible in the health discourse surrounding \textit{gomutra} (cow urine) in India. While recent literature highlights the spread of pseudoscience, narrative-based remedies, and the zoonotic risks associated with \textit{gomutra} consumption during health crises like the COVID-19 pandemic \cite{Essar2021CowDungCOVID, Hurford2022NarrativeMisinformationIndia}, this is precisely the kind of culturally embedded, non-Western discourse that existing computational approaches are least equipped to handle, and yet it remains entirely uncharted without any datasets, linguistic analyses for this specific discourse. Furthermore, there are no existing works on LLM-based evaluations of the rhetorical strategies driving \textit{gomutra}-related health misinformation in Indian languages, and no multilingual datasets exist in this regard.

This study addresses these critical gaps by introducing the first multilingual dataset on cow-urine discourse and providing a culturally aware, LLM-assisted discourse analysis of these traditional narratives, including sensitivity to prompting. 


\section{Methodology and Evaluation}
Below, we present the methodology for data collection for our preliminary experiments and evaluation of the LLMs.


\subsection{Dataset Collection and Curation}

We collected YouTube videos related to \textit{gomutra} discourse using the YouTube Data API v3. To capture relevance and linguistic variation, we queried the API using a combination of keywords and hashtags, including \textit{gomutra}, cow urine, cowurine, \textit{gaumutra}, \textit{gaumutra} health, \textit{gomutra} health, \textit{\#gomutra}, \#gomata, and \#cowurine. The search query returned videos in multiple languages (e.g., English, Hindi, Urdu, Kannada, Tamil, and Gujarati), which were curated via a two-step process. First, titles, descriptions, and audio were manually reviewed to ensure topic relevance, retaining only videos with primary spoken languages of English, Hindi, or Urdu. Second, after removing duplicates and derivative content, we relied on the API's default search relevance to finalize our curated subset of 30 videos for this preliminary exploratory study.

For each curated video, we extracted the audio track using the open-source yt-dlp library. To ensure strict compliance with the platform's terms of service, we do not store or redistribute raw audio files; all downstream processing relies exclusively on publicly accessible content for non-commercial research. Audio was transcribed using OpenAI's Whisper model (the large checkpoint) \cite{radford2022robust}, selected for its state-of-the-art accuracy across multiple languages. 

To evaluate transcription accuracy, we calculated the Word Error Rate (WER) on a randomly selected subset comprising 16\% of the video corpus. A single author manually reviewed the machine-generated text against the human-generated reference transcript to compute the WER using the following standard formulation:

$$\mathrm{WER} = \frac{S + D + I}{N}$$

where $S$ represents the number of substitutions, $D$ is the number of deletions, $I$ is the number of insertions, and $N$ is the total number of words in the reference transcript. The average WER across this sampled dataset was 7.04\%, demonstrating a high level of transcription fidelity for our analysis. These non-English transcripts were translated into English using GPT-4 \cite{openai2023gpt4} via the OpenAI API. Full details regarding transcription hyperparameters and translation prompt design are provided in the Appendix.

Finally, utilizing predefined written guidelines, a single author manually annotated each video to determine both the speaker’s gender and the overall stance toward \textit{gomutra} (promoting or debunking), thereby establishing the ground truth for our comparative analysis. We structured our gender annotations to reflect the varied formats of these videos, clearly differentiating between monologues with a single primary speaker and interactive, discussion-style videos featuring multiple speakers. The resulting dataset consists of $N=30$ unique videos. Analysis at the video level reveals a highly skewed stance distribution: 24 videos (80\%) promote the use of \textit{gomutra}, while 6 (20\%) debunk it. To account for videos featuring interactive, multi-speaker formats, our gender analysis was conducted at the speaker level ($N_{speakers}=33$). This also yields a skewed demographic breakdown of 20 male (60.6\%) and 13 female (39.4\%) speakers. All computational experiments and data processing were conducted using Google Colaboratory. Detailed hardware specifications and compute budget information are provided in the Appendix. 

To provide concrete examples of the health narratives present in our dataset, we highlight a few translated excerpts from the transcripts. Many videos frame the substance as a comprehensive cure for lifestyle diseases, making broad claims such as: \textit{``As a way of life, those who are overweight should drink cow urine... It is a permanent solution to constipation, a medicine for obesity, it purifies the blood, and detoxifies the liver, kidneys, and the entire body.''} Other videos attempt to manufacture medical credibility by invoking institutional authority, stating: \textit{``A patent has already been obtained for using cow urine to manufacture medicines for cancer and other diseases, and currently, the patent certificate is displayed on your screen.''} Furthermore, transcripts frequently target specific ailments by blending traditional terminology with modern medical jargon, claiming that \textit{ ``because it reduces phlegm and wind, people suffering from high cholesterol, fatty liver, or blockages in the body, low bone density, or more bloating, all of them can consume it.''}

\subsection{Experimental Setup and LLM Evaluation}
Multiple Large Language Models were evaluated for their ability to interpret these competing narratives. First, to analyze macro-level discourse, GPT-4o was employed using a standardized prompt to identify two primary linguistic features: (i) traditional metaphors, defined as narratives deeply connected to cultural heritage (e.g., Ayurvedic medicines, religious references like divine or amrit) to establish credibility; and (ii) scientific terms, comprising pseudo-scientific language that emphasizes empirical evidence and clinical terminology (e.g., antioxidants, toxins). Term densities for these identified features were computed per 100 words across both promoting and debunking stances. To quantify the GPT-4o performance against the human-annotated ground truth, we computed precision ($P$), recall ($R$), and the F1-score ($F1$). These metrics are based on the identification of True Positives ($TP$), False Positives ($FP$), and False Negatives ($FN$). Specifically for this study, a true positive is defined as an extracted scientific or traditional term that exists within the set of human-annotated ground-truth terms by a native speaker. These evaluation metrics were calculated for each transcript utilizing the standard mathematical formulations described in the Appendix. The final evaluation scores reported in the main text represent the macro-average of these metrics across all transcripts in the manually annotated validation subset.

Second, to operationalize a theoretically grounded linguistic marker of persuasive intent that is both computationally tractable and linguistically interpretable, intensifiers were selected as a focal linguistic feature because of their established roles in persuasive and deceptive communication \cite{CHEUNG2025101492,  HOLMES1990185}. An intensifier is a word or a phrase specifically used to strengthen, exaggerate, or emphasize a claim. It is a lexical item that operates as a degree modifier on an adjective, an adverb, or occasionally a verb
phrase. For example, \textit{very} (as in "The treatment is very effective"), \textit{completely} (as in "It completely cures the disease"), and \textit{absolutely} (as in "This is absolutely proven that cow urine has anti-cancer properties"). Unlike neutral descriptors, intensifiers signal epistemic uncertainty, a well-documented characteristic of health misinformation, which tends toward absolute claims rather than the hedged language typical of evidence-based communication \cite{VanDerLinden2022Misinformation}. By quantifying intensifier use across promotional and debunking content, we evaluated the following models: GPT-4o-mini, Gemini 2.5 Pro, and DeepSeek-V3.1. To ensure accurate computation, repeated intensifiers within a single model output were filtered out to yield unique counts. For this intensifier analysis, we designed a structured prompting strategy to evaluate model sensitivities, comparing zero-shot (no examples provided) against structured few-shot environments. Furthermore, motivated by prior work demonstrating that surface-level stylistic variations in prompts can significantly affect LLM outputs \cite{guan, butterfly, sclar2024, zhuo-etal-2024-prosa}, we investigated the impact of prompt personas by comparing a formal instruction style against a "friendly" conversational tone (e.g., "Hi LLM!"), to analyze whether social cues encourage the models to produce more nuanced, human-like language extraction (full prompts are provided in the Appendix). In addition, to assess cross-model reliability of intensifier identification, we computed pairwise Cohen's Kappa ($\kappa$) \cite{cohen1960, landis1977} on a presence/absence matrix of the unique intensifiers identified across all conditions. Cohen's Kappa was selected over simple percentage agreement because it corrects for chance agreement, making it appropriate for comparing binary identification decisions across models with different output volumes. For each model pair, a vector of 1s and 0s was constructed indicating whether each intensifier was flagged by that model in any of its four conditions; Kappa was then computed on these vectors. Condition-level Kappa was additionally computed by collapsing across models to compare formal versus friendly tone conditions and zero-shot versus few-shot settings independently.

Complete code is available at Github repository: 
\url{https://github.com/MsAnamtaKhan/gomutra_health_misinformation}


\section{Core Findings}

\begin{figure}[t]
  \centering
  
   \includegraphics[width=0.8\linewidth]{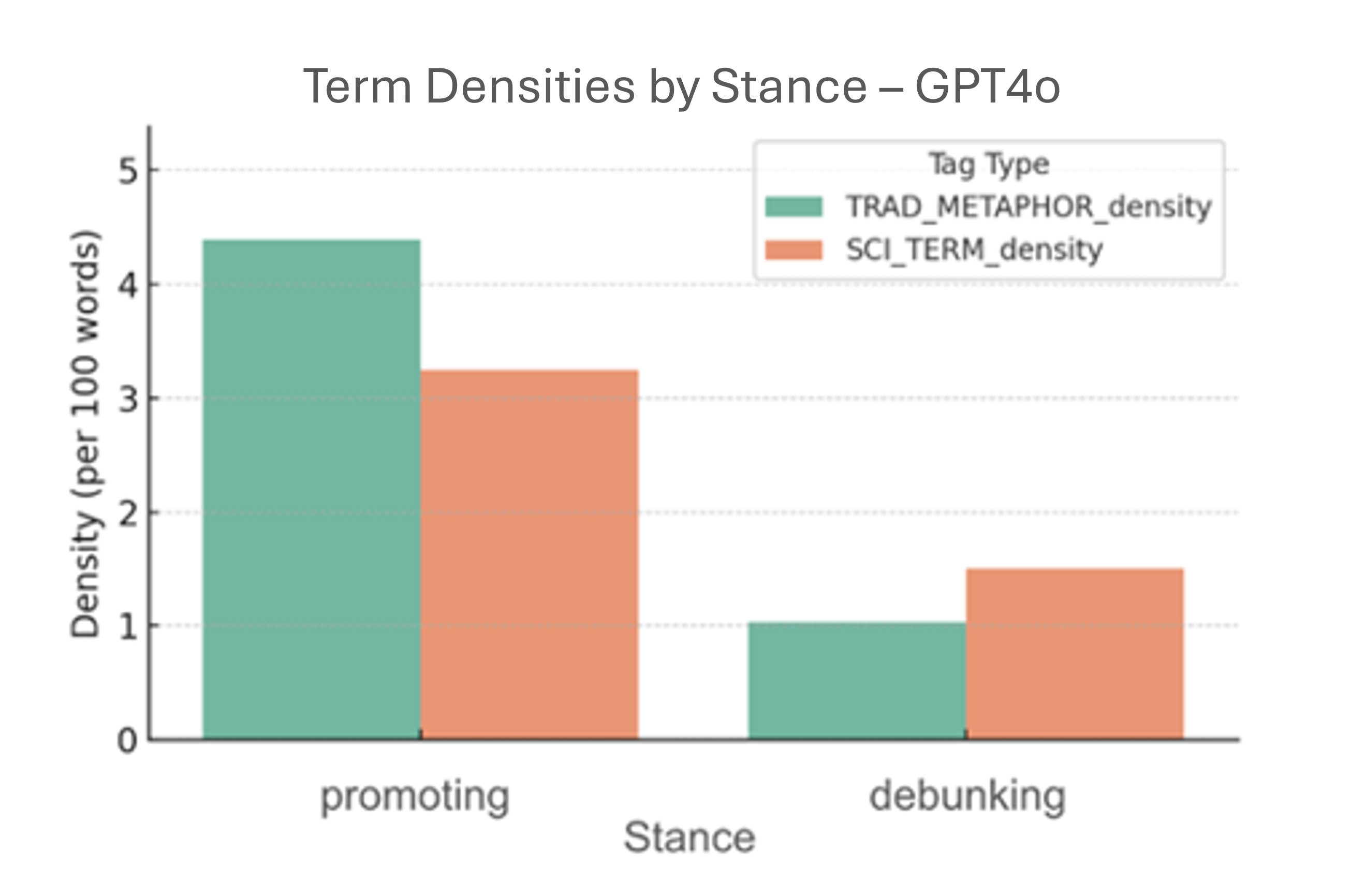}

   \caption{Comparison of traditional metaphor and scientific term densities (per 100 words)  given by GPT-4o  across promoting and debunking stances.}
   \label{fig:densities-graph}
\end{figure}

Our analysis addresses each of the three research questions posed in the introduction and yields three principal findings:

\textbf{\textit{RQ1: Asymmetric Rhetorical Framing:}}
Our analysis revealed a significant asymmetry in linguistic strategy between promotional and debunking \textit{gomutra} content.  We found that to establish credibility,  the speakers leveraged scientific terms (e.g., \textit{betadine}, \textit{detoxifies}, \textit{immunity}, \textit{arthritis}) to project medical legitimacy, while simultaneously employing traditional metaphors (e.g., \textit{Ayurveda}, \textit{amrit}, \textit{sanjeevani}) to anchor their claims in a deep cultural context.  As shown in Figure \ref{fig:densities-graph}, promotional content exhibited substantially higher densities of both traditional metaphors ($\sim$ 4.4 per 100 words) and scientific terms ($\sim$ 3.2 per 100 words), with traditional metaphors dominating. Debunking content, by contrast, showed markedly lower densities overall ($\sim$1.0 traditional metaphors and $\sim$1.5 scientific terms per 100 words), but crucially employed a more \textit{balanced} rhetorical approach, integrating both scientific terminology and traditional references rather than relying on either exclusively. This balance was statistically confirmed: promotional (positive) content showed significantly higher density of traditional terms (t = 2.61, p $<$ 0.05; Mann-Whitney U = 130.5; p $<$ 0.05), while debunking (negative) content's more even distribution was similarly significant (t = 2.63, p $<$ 0.05; Mann-Whitney U = 133.0; p $<$ 0.05) \cite{macfarland2016mann}. To evaluate the performance of the LLM in extracting scientific and traditional terms, we randomly selected one-third of the dataset, consisting of both promoting and debunking content, for manual annotation by a single author (a native speaker of Hindi). We calculated the precision, recall, and F1-score, as shown in Table\ref{tab:rq1_prf}, by comparing the human ground-truth annotations against the GPT-4o outputs for each transcript, reporting the average across the subset. For scientific terms, the model achieved an average precision of 61\%, a recall of 53\%, and an F1-score of 52\%. For traditional terms, it achieved a precision of 64\%, a recall of 56\%, and an F1-score of 59\%. These results demonstrate that our automated extraction pipeline provides a sufficiently robust and reliable baseline for evaluating the broader dataset in this exploratory study.

\begin{table}[h]
\centering

\renewcommand{\arraystretch}{1.2}
\setlength{\tabcolsep}{5pt}
\begin{tabular}{lccc}
\hline
\textbf{Linguistic Strategy} & \textbf{Precision} & \textbf{Recall} & \textbf{F1-Score} \\
                            & \textbf{(\%)}      & \textbf{(\%)}   & \textbf{(\%)} \\
\hline
Scientific terms  & 61 & 53 & 52 \\
Traditional terms & 64 & 56 & 59 \\
\hline
\end{tabular}
\caption{Evaluation of GPT-4o term extraction performance against human annotation.}
\label{tab:rq1_prf}

\end{table}
\footnotetext{Note that GPT-4 and GPT-4o were employed for transcript translation and intensifier/prompt-sensitive analysis, respectively.}

This asymmetry creates a direct challenge for LLM-based detection. The finding that debunking content employs traditional terminology reflects a communicatively rational strategy: to persuade an audience already invested in traditional medicine, effective debunkers must engage the cultural framework on its own terms rather than simply opposing it with scientific language, which risks being dismissed as culturally alien or elitist. This mirrors well-documented patterns in science communication, where meeting audiences within their existing belief frameworks is more persuasive than direct contradiction \cite{VanDerLinden2025HealthMisinformation}. A model that treats high traditional metaphor density as a marker of misinformation will therefore perform well on promotional content, but will systematically misclassify this sophisticated debunking strategy, flagging culturally competent counter-speech as the very misinformation it is trying to correct.

\textbf{\textit{RQ2: “Friendly” Prompt Doesn’t Always Mean LLMs are More Expressive}}
Contrary to our hypothesis, varying prompt persona from formal to friendly (see Appendix), zero-shot to few-shot (i.e., providing a few examples) produced no consistent difference in LLM output quality across models. Table \ref{tab:intensifiers_comparison} illustrates this inconsistency clearly: the effect of the friendly persona differed in both direction and magnitude across models. Gemini 2.5 Pro showed only a marginal decrease from formal to friendly in zero-shot (21 vs 45 unique intensifiers respectively), while GPT 4o-mini showed a dramatic drop under the friendly condition (48 formal vs 34 few-shot formal, and 14 friendly vs 24 few-shot friendly). DeepSeek-V3.1, on the other hand showed the opposite pattern in few-shot settings, producing slightly more intensifiers under the friendly condition (13 formal vs. 16 friendly). Similarly, few-shot prompting did not consistently outperform zero-shot across models: for GPT-40-mini, few-shot reduced intensifier counts under the formal condition, while for DeepSeek-V3.1, it increased them under the friendly condition. 

Inter-model agreement was consistently poor across all pairs 
(Table~\ref{tab:kappa}). Gemini and GPT-4o-mini showed below-chance 
agreement ($\kappa = -0.256$), indicating systematic divergence rather 
than random disagreement. Gemini and DeepSeek-V3.1 showed near-zero agreement 
($\kappa = -0.006$), suggesting effectively independent identification. 
GPT-4o-mini and DeepSeek-V3.1 showed the highest agreement, though still only 
fair ($\kappa = 0.214$). Condition-level analysis revealed comparable 
divergence: formal versus friendly prompting produced $\kappa = -0.431$, 
and zero-shot versus few-shot produced $\kappa = -0.452$, indicating that 
both tone and shot setting caused models to identify systematically 
different intensifiers rather than merely different quantities.

\begin{table}[]
\centering
\begin{tabular}{lcc}
\hline
\textbf{Comparison} & \textbf{Cohen's $\kappa$}  \\
\hline
Gemini vs GPT-4o-mini    & $-0.256$  \\
Gemini vs DeepSeek-V3.1       & $-0.006$ \\
GPT-4o-mini vs DeepSeek-V3.1  & $0.214$  \\
Formal vs Friendly       & $-0.431$  \\
Zero-shot vs Few-shot    & $-0.452$ \\
\hline
\end{tabular}
\caption{Pairwise Cohen's $\kappa$ across models and prompt conditions.}
\label{tab:kappa}
\end{table}

This inconsistency is itself a substantive finding: LLMs do not respond to social register cues the way human interlocutors do, and cultural competency in automated analysis cannot be achieved through surface-level persona adjustment alone. Deeper architectural and training-level solutions are required in future studies.

\textbf{\textit{RQ3: Speaker Specific Rhetorical Patterns and Gender Bias Risk}}
We identified a distinct gender-based pattern in their rhetorical strategy across the corpus, though we note that the dataset is skewed toward male speakers, reflecting the broader gender imbalance in \textit{gomutra} promotional content on YouTube. As detailed in Table \ref{tab:rq3_stats},  male speakers favored a rhetoric of certainty and exaggeration. They deployed absolute-certainty amplifiers (“never”, “permanent”) at a distinctly higher rate than their female counterparts (averaging 5.64 occurrences per 1,000 words versus just 0.67). Furthermore, men utilized universality markers (“all,” “completely,” “everywhere”) more frequently (11.51 vs. 9.04 per 1,000 words) and relied heavily on authoritative, first-person framing (“I”), which appeared an average of 10.27 times per 1,000 words for males compared to only 2.61 for females. Notably, the use of hyperbolic terms (“immortal,” “miracle”) was exclusively observed in male speech within this sample, averaging 1.99 instances per 1,000 words. Female speakers, by contrast, employed relational and inclusive language (“we,” “our body”) paired with gentle encouragement, building trust through community rather than top-down authority. This shift toward communal framing is quantitatively reflected in their higher frequency of inclusive pronouns ("we/our"), which appeared 8.94 times per 1,000 words in female speech compared to 7.89 in male speech, standing in stark contrast to the male preference for the singular "I."

\begin{table}[htbp]
\centering

\begin{tabular}{llcc}
\toprule
& & \multicolumn{2}{c}{\textbf{Shot Setting}} \\
\cmidrule(lr){3-4}
\textbf{Model} & \textbf{Prompt Style} & \textbf{Zero-shot} & \textbf{Few-shot} \\
\midrule
Gemini & Formal & 21 & 35 \\
       & Friendly & 45 & 44 \\
GPT4o-mini & Formal & 48 & 34 \\
           & Friendly & 14 & 24 \\
DeepSeek-V3.1 & Formal & 13 & 16 \\
         & Friendly & 12 & 9 \\
\bottomrule
\end{tabular}
\caption{Comparison of the Number of Unique Intensifiers Generated across Different Models (Gemini, GPT4o-mini, and DeepSeek-V3.1) and Prompt Styles (Formal vs. Friendly), Contrasting Zero-shot and Few-shot Settings.}

\label{tab:intensifiers_comparison}
\end{table}

These patterns are consistent with broader findings in gender and communication research \cite{lakoff1973language, deFrancisco1992, HOLMES1990185}. Contemporary studies confirm that these dynamics persist in modern digital spaces; for instance, recent work highlights that female social media users continue to employ more indirect, affiliative language such as hedging \cite{Aydin2025GenderDifferences}, whereas male digital influencers tend to rely heavily on direct, assertive framing \cite{WangChen2022GenderSocialMedia}. These gendered communication styles have direct implications for automated moderation.  LLMs trained without awareness of gendered rhetorical norms risk systematically misreading female-coded persuasion, which avoids absolutes and relies on relational framing, as less confident or less promotional than it actually is. This introduces a structural gender bias into moderation outcomes that is not a minor calibration issue but a fairness concern with real deployment consequences.

\begin{table*}[t]
\centering
\begin{tabular}{lcc}
\toprule
& \multicolumn{2}{c}{\textbf{Mean per 1000 Words}} \\
\cmidrule(lr){2-3}
\textbf{Linguistic Markers} & \textbf{Male} & \textbf{Female} \\
\midrule
Certainty amplifiers ("never", "permanent")   & 5.64 & 0.67 \\
Universality markers ("all", "completely") & 11.51 & 9.04  \\
First-person "I" & 10.27 & 2.61  \\
Inclusive "we/our"  & 7.89 & 8.94  \\
Hyperbolic terms & 1.99 & 0  \\
\bottomrule
\end{tabular}
\caption{Comparison of linguistic markers  between male and female speakers, measured as a mean per 1,000 words.}
\label{tab:rq3_stats}
\end{table*}



\section{Implications }

Our findings do not resolve how LLMs should be used to analyse culturally embedded health misinformation; they reveal the conditions under which such analysis would be unreliable, and what would need to change. There are three main implications of our work from this perspective. First, we reconfirm a finding that has been much talked about in scholarship and the mainstream media alike, that LLMs fail to parse rhetorical signals specific to cultures outside of those on whose data they are built upon. This, in effect means they fail to capture nuance in culturally embedded misinformation. This is articulated in the finding that an understanding of tradition-specific metaphor and its employment is essential in comparing the discursive approaches of promoters and debunkers of gomutra-related theories. Second, we find that even slight alterations in the nature of conversations with LLMs has sizable impacts on the prompt responses. Finally, we find that there are unique elements of gendered behavior that can additionally undermine LLMs' ability to offer fair analyses and outputs of complex narrative content.

The proliferation of such content presents three intertwined challenges: \textit{Epistemically,} culturally embedded misinformation resists verification because it operates through belief systems rather than factual falsehood. This is observed through both the invocation of identity through the call to tradition, but also in the subtle ways in which patriarchy manifests itself subtly in distinctions between male and female influencers' output of the same narrative. \textit{Infrastructurally,} platform algorithms amplify this content even before analyses can occur \cite{chen2024combating, tao2024cultural}. That both debunkers and promoters use a mix of science and tradition underlines the perceived value for rhetorical inversion among influencers, who as sophisticated parties in the information economy, understand affective values attached to the cases they wish to make. The hybrid epistemic registers appeal to different elements in the social network to push virality. \textit{Rhetorically,} promotional content exploits cultural familiarity and traditional authority in ways that differ fundamentally from straightforwardly false claims, and that LLMs trained predominantly on Western corpora are structurally ill-equipped to recognise. Here, we see that there is use of obvious linguistic cues, that machines can be trained to recognize, but also less obvious metaphorical references that require an understanding of cultural elements specific to India to accurately debunk misinformation.


\textbf{The Benchmark Problem} Existing misinformation benchmarks are built predominantly on Western, English-language datasets where rhetorical norms, cultural references, and authority structures differ substantially from the Indian context studied here. Our results show that this gap is not merely a matter of translation: LLMs systematically misread the rhetorical inversions specific to this domain, where debunking content \textit{engages} with traditional narratives rather than simply opposing them. A model trained to treat scientific language as a marker of credibility will misread sophisticated debunking content that strategically deploys cultural references for persuasive effect, mistaking culturally competent counter-speech for the very misinformation it is trying to counter. 

\textbf{The case for Indic LLMs} Our work here provides a persuasive case for regionally trained language models for fine-grained analysis of layered discourse. Western general-purpose LLMs struggle with sarcasm, cultural idiom, and code-switching patterns characteristic of multilingual Indian health discourse - features that are not edge cases but structural properties of how this content circulates \cite{gumma2024healthpariksha, compositeharm2026}. What we see here goes beyond the understanding of words and terms to a deeper meaning of what is being said. Since innuendo and figures of speech are commonly found in an misinformation, it makes for a useful case study, but arguably a range of domains in casual or even literary conversation have affective elements that will be lost without culturally thoughtful human engagement.   

\textbf{Do LLMs have a gender problem?}
The gendered rhetorical patterns we identify have direct implications for LLM-assisted discourse analysis. A model that reads absolute-certainty language as the primary marker of promotional content will systematically underread female-coded persuasion, which operates through relational framing and inclusive language rather than hyperbole. These are not minor calibration issues but structural analytical blind spots that risk misrepresenting whose voices get flagged and whose do not. While our work offers more direct insight into gendered style creeping into social media commentary, a larger question remains whether there is a case for more work into how LLMs understand a world in which the overwhelming majority of conversations it is built on are in the voices of men.

These findings are preliminary and derived from a small, culturally specific corpus (a small-scale analysis). We do not claim direct generalizability to other traditional medicine misinformation domains, though we believe our work, examining asymmetric rhetorical strategies, testing prompt sensitivity, and auditing for gendered patterns, is replicable across culturally diverse discourse contexts. What this study demonstrates is that LLM-assisted discourse analysis in non-Western contexts requires cultural grounding that goes beyond translation and prompt design, it requires rethinking what counts as a reliable analytical instrument in the first place.








\section{Limitations}

This study has several limitations that future work should address. First, the corpus of 30 transcripts, while carefully selected, limits statistical power and generalizability; findings should be treated as hypothesis-generating rather than definitive. Second, our reliance on a multilingual translation model introduces potential noise, particularly for code-switched or dialect-specific content. Translation quality was not explicitly evaluated, and models trained exclusively on Indian languages may perform differently, potentially affecting downstream applications. Third, we probe only three LLMs and do not compare against traditional supervised classifiers, which would provide a more complete picture of the detection landscape. Fourth, stance (promoting/debunking) and speaker's gender were annotated by a single author. Finally, the retirement of GPT-4o mini from OpenAI's consumer interface during the study period prevented a fully controlled comparison across all conditions. These limitations define a clear agenda for future work: a larger, human-annotated corpus; direct comparison with Indic LLMs and traditional classifiers; and extension to related traditional medicine misinformation domains in other cultural contexts. Additionally, all prompts were administered in English to Western-developed LLMs. Future work should explore prompting in Indian languages and providing original (untranslated) transcripts directly to the model, which may yield culturally closer responses.


\section{Conclusion}
This paper examined how culturally embedded health misinformation - specifically, promotional and debunking content around \textit{gomutra} on Indian YouTube - differs rhetorically, using LLMs as analytical instruments for a post-facto examination of discourse. Our analysis reveals an asymmetry: promotional content relies more heavily on traditional metaphor and cultural authority, while effective debunking integrates both scientific and traditional framing to build credibility. Yet, we do find that promotional content also uses science and scientific terms to emphasize empirical credibility. LLMs, we find, are sensitive to these differences in ways that are not yet well-controlled, responding inconsistently to prompt tone, misreading gendered rhetorical strategies, and struggling with the cultural inversions that characterize sophisticated debunking. These findings have implications beyond \textit{gomutra}. As LLMs are increasingly proposed as scalable moderation tools across languages and culture, this study illustrates that cultural competency cannot be retrofitted through prompt engineering alone; it requires training data, benchmarks, and evaluation frameworks that reflect the rhetorical diversity of the global information environment. 

Culturally embedded health misinformation on YouTube does not look like ordinary misinformation; it blends sacred traditional language with pseudo-scientific claims in ways that even sophisticated debunking content mirrors, and LLMs used to analyse this discourse are systematically misled by these rhetorical patterns in ways that differ by culture, gender, and prompt design.

In the future, we plan to expand this work to larger corpora and also compare the performances of Western models with Indic language models to prevent the loss of cultural nuance inherent in translation, thereby allowing us to strictly preserve Hindi-English code-switching and capture richer sociolinguistic signals \cite{coli_veridicality,perez2025detecting}. Additionally, we also plan to broaden our scope beyond textual analysis to incorporate multimodal analyses, including the full visual context of the videos \cite{shang2025multitec}; leveraging Optical Character Recognition (OCR) to extract on-screen text and identifying visual cues, such as clinical props, medical reports, and affiliate links, to better map the use of authority and financial incentives driving these claims. Furthermore, we also intend to benchmark current models against a more comprehensive suite of baselines, including classical machine learning classifiers, specialized Indic LLMs \cite{danish2026comparative}, and foundational linguistic metrics such as lexicon, part-of-speech (POS), and dependency-based structural counts.





\section{Ethical considerations}
Our research touches upon deeply rooted cultural beliefs surrounding \textit{gomutra}. To navigate this respectfully, our analysis focuses strictly on the rhetorical strategies that influencers use to spread unverified health claims, rather than presenting a critque of the cultural or traditional practices that underlie the willingness to think of \textit{gomutra} as medicinal. In this, our goal is not to critique traditional medicine more broadly, but rather to critically examine public-facing actors and the algorithmic systems that amplify these narratives.  All data was collected exclusively from publicly available YouTube videos. We intentionally avoided collecting Personally Identifiable Information (PII) from viewers or commenters. To comply with platform guidelines, our aggregated dataset will be released strictly for non-commercial, academic research under an ethical use agreement.

Finally, we acknowledge the inherent fairness risks in our gender-based analysis. Annotating speaker gender as a binary variable (male/female) based on visual presentation is deconstructive; it fails to capture the full spectrum of gender identity and carries the risk of misgendering. This methodological compromise was, however, necessitated by the lack of self-reported demographic data. We emphasize that this categorization was used solely as a diagnostic lens to analyze linguistic patterns and expose gender-based biases in LLM outcomes, rather than to essentialize communication styles or reinforce restrictive gender binaries.

\section{Acknowledgements}
We would like to thank all the reviewers for their valuable comments and feedback. We are also grateful to Ananya Sharedalal for proposing the idea of investigating cow urine and for compiling the initial list.

\bibliography{aaai2026}

@article{guan,
  title={The Order Effect: Investigating Prompt Sensitivity to Input Order in LLMs},
  author={Guan, Bryan and Roosta, Tanya and Passban, Peyman and Rezagholizadeh, Mehdi},
  journal={arXiv},
  volume={abs/2502.04134},
  year={2025}
}

@inproceedings{zhuo-etal-2024-prosa,
    title = "{P}ro{SA}: Assessing and Understanding the Prompt Sensitivity of {LLM}s",
    author = "Zhuo, Jingming  and
      Zhang, Songyang  and
      Fang, Xinyu  and
      Duan, Haodong  and
      Lin, Dahua  and
      Chen, Kai",
    editor = "Al-Onaizan, Yaser  and
      Bansal, Mohit  and
      Chen, Yun-Nung",
    booktitle = "Findings of the Association for Computational Linguistics: EMNLP 2024",
    month = nov,
    year = "2024",
    address = "Miami, Florida, USA",
    publisher = "Association for Computational Linguistics",
    url = "https://aclanthology.org/2024.findings-emnlp.108/",
    doi = "10.18653/v1/2024.findings-emnlp.108",
    pages = "1950--1976"
}

@inproceedings{butterfly,
    title = "The Butterfly Effect of Altering Prompts: How Small Changes and Jailbreaks Affect Large Language Model Performance",
    author = "Salinas, Abel  and
      Morstatter, Fred",
    editor = "Ku, Lun-Wei  and
      Martins, Andre  and
      Srikumar, Vivek",
    booktitle = "Findings of the Association for Computational Linguistics: ACL 2024",
    month = aug,
    year = "2024",
    address = "Bangkok, Thailand",
    publisher = "Association for Computational Linguistics",
    url = "https://aclanthology.org/2024.findings-acl.275/",
    doi = "10.18653/v1/2024.findings-acl.275",
    pages = "4629--4651"
}

@inproceedings{sclar2024,
  title={Quantifying Language Models' Sensitivity to Spurious Features in Prompt Design or: How I learned to start worrying about prompt formatting},
  author={Sclar, Melanie and Choi, Yejin and Tsvetkov, Yulia and Suhr, Alane},
  booktitle={The Twelfth International Conference on Learning Representations (ICLR)},
  year={2024},
  url={https://arxiv.org/abs/2310.11324}
}

@article{gumma2024healthpariksha,
  author    = {Gumma, Varun and Raghunath, Ananditha 
               and Jain, Mohit and Sitaram, Sunayana},
  title     = {{HEALTH-PARIKSHA}: Assessing {RAG} Models 
               for Health Chatbots in Real-World 
               Multilingual Settings},
  journal   = {arXiv preprint arXiv:2410.13671},
  year      = {2024},
  url       = {https://arxiv.org/abs/2410.13671}
}

@article{compositeharm2026,
  title={Lost in Translation? A Comparative Study on the Cross-Lingual Transfer of Composite Harms},
  author={Shukla, Vaibhav and Sharma, Hardik and Reganti, Adith N and Wasmatkar, Soham and Kumar, Bagesh and Singh, Vrijendra},
  journal={arXiv preprint arXiv:2602.07963},
  year={2026}
}

@article{CHEUNG2025101492,
title = {Use of Linguistic Communication Strategies (Hedges and Intensifiers) in Simulated Pharmacy Education Shared Decision-Making},
journal = {American Journal of Pharmaceutical Education},
volume = {89},
number = {10},
pages = {101492},
year = {2025},
issn = {0002-9459},
doi = {https://doi.org/10.1016/j.ajpe.2025.101492},
url = {https://www.sciencedirect.com/science/article/pii/S0002945925001378},
author = {Natalie Cheung and Averil Grieve and Kyle Wilby and Tim Tran and Angelina Lim}
}

@article{landis1977,
  author    = {Landis, J. Richard and Koch, Gary G.},
  title     = {The Measurement of Observer Agreement for Categorical Data},
  journal   = {Biometrics},
  year      = {1977},
  volume    = {33},
  number    = {1},
  pages     = {159--174},
  doi       = {10.2307/2529310}
}

@article{cohen1960,
  title={A coefficient of agreement for nominal scales},
  author={Cohen, Jacob},
  journal={Educational and psychological measurement},
  volume={20},
  number={1},
  pages={37--46},
  year={1960},
  publisher={Sage Publications Sage CA: Thousand Oaks, CA}
}

@inproceedings{jin2024better,
  author    = {Jin, Yiqiao and Chandra, Mohit and Verma, Gaurav and Hu, Yibo and De Choudhury, Munmun and Kumar, Srijan},
  title     = {Better to Ask in English: Cross-Lingual Evaluation of Large Language Models for Healthcare Queries},
  booktitle = {Proceedings of the ACM Web Conference 2024},
  series    = {WWW '24},
  year      = {2024},
  pages     = {2627--2638},
  doi       = {10.1145/3589334.3645643},
  url       = {https://doi.org/10.1145/3589334.3645643},
  publisher = {Association for Computing Machinery}
}

@article{bedi2025testing,
  title={Testing and evaluation of health care applications of large language models: a systematic review},
  author={Bedi, Suhana and Liu, Yutong and Orr-Ewing, Lucy and Dash, Dev and Koyejo, Sanmi and Callahan, Alison and Fries, Jason A and Wornow, Michael and Swaminathan, Akshay and Lehmann, Lisa Soleymani and others},
  journal={Jama},
  volume={333},
  number={4},
  pages={319--328},
  year={2025}
}

@misc{indiatoday2022choubeydesai,
  author       = {{Asian News International}},
  title        = {Former PM Morarji Desai also used to drink cow urine for medicinal benefits: Ashwini Choubey},
  year         = {2022},
  month        = feb,
  day          = {4},
  journal      = {India Today},
  howpublished = {News article},
  address      = {Patna},
  url          = {https://www.indiatoday.in/},
  note         = {Updated February 4, 2022}
}

@misc{reuters2020trumpdisinfectant,
  author       = {Kelland, Kate and Satter, Raphael},
  title        = {Trump's COVID-19 Disinfectant Ideas Horrify Health Experts},
  year         = {2020},
  month        = apr,
  day          = {24},
  journal      = {Reuters},
  howpublished = {News article},
  url          = {https://www.reuters.com/world/trumps-covid-19-disinfectant-ideas-horrify-health-experts/},
  note         = {Updated April 26, 2020}
}

@misc{scio2020fighting,
  author       = {{State Council Information Office of the People's Republic of China}},
  title        = {Fighting COVID-19: China in Action},
  year         = {2020},
  month        = jun,
  institution  = {State Council Information Office of the People's Republic of China},
  howpublished = {White paper}
}

@misc{reuters2024ramdevmisled,
  author       = {Chaturvedi, Arpan},
  title        = {Indian State Says Yoga Guru Misled Public with COVID, Other Cures},
  year         = {2024},
  month        = may,
  day          = {7},
  journal      = {Reuters},
  howpublished = {News article},
  url          = {https://www.reuters.com/world/india/indian-state-says-yoga-guru-misled-public-with-covid-other-cures-2024-05-07/}
}

@misc{indiatoday2019cowurine,
  author       = {{India Today}},
  title        = {Cow Urine to Be Used for Preparing Medicines, Treating Cancer: Health Minister Ashwini Kumar Choubey},
  year         = {2019},
  month        = sep,
  day          = {8},
  journal      = {India Today},
  howpublished = {News article},
  url          = {https://www.indiatoday.in/india/story/former-pm-morarji-desai-also-used-to-drink-cow-urine-for-medicinal-benefits-ashwini-choubey-1596840-2019-09-08},
  note         = {Published September 8, 2019}
}

@misc{kff2024aihealth,
  author       = {{Kaiser Family Foundation}},
  title        = {Poll: Most Who Use Artificial Intelligence Doubt AI Chatbots Provide Accurate Health Information},
  year         = {2024},
  month        = aug,
  day          = {15},
  howpublished = {KFF News Release},
  institution  = {Kaiser Family Foundation},
  url          = {https://www.kff.org/health-information-trust/poll-most-who-use-artificial-intelligence-doubt-ai-chatbots-provide-accurate-health-information/}
}

@article{garg2025context,
  author        = {Garg, Prashant and Fetzer, Thiemo},
  title         = {How Much Does Context Affect the Accuracy of AI Health Advice?},
  journal       = {arXiv preprint arXiv:2504.18310},
  year          = {2025},
  eprint        = {2504.18310},
  archiveprefix = {arXiv},
  primaryclass  = {econ.GN},
  doi           = {10.48550/arXiv.2504.18310},
  url           = {https://doi.org/10.48550/arXiv.2504.18310},
  note          = {Version 2, revised February 24, 2026}
}

@article{khalil2025evaluating,
  author  = {Khalil, Mahmoud and Mohamed, Fatma and Shoufan, Abdulhadi},
  title   = {Evaluating the Quality of Medical Content on YouTube Using Large Language Models},
  journal = {Scientific Reports},
  year    = {2025},
  month   = mar,
  day     = {22},
  volume  = {15},
  number  = {1},
  pages   = {9906},
  doi     = {10.1038/s41598-025-94208-6},
  pmid    = {40121315},
  pmcid   = {PMC11929840}
}

@article{kong2021tiktok,
  author  = {Kong, Wei and Song, Shuang and Zhao, Yifan and Zhu, Qi and Sha, Libo},
  title   = {TikTok as a Health Information Source: Assessment of the Quality of Information in Diabetes-Related Videos},
  journal = {Journal of Medical Internet Research},
  year    = {2021},
  volume  = {23},
  number  = {9},
  pages   = {e30409},
  doi     = {10.2196/30409},
  url     = {https://www.jmir.org/2021/9/e30409}
}

@article{li2022youtube,
  author  = {Li, Heidi Oi-Yee and Pastukhova, Elena and Brandts-Longtin, Olivier and Tan, Marcus G. and Kirchhof, Mark G.},
  title   = {YouTube as a Source of Misinformation on COVID-19 Vaccination: A Systematic Analysis},
  journal = {BMJ Global Health},
  year    = {2022},
  month   = mar,
  day     = {9},
  volume  = {7},
  number  = {3},
  pages   = {e008334},
  doi     = {10.1136/bmjgh-2021-008334},
  pmid    = {35264318},
  pmcid   = {PMC8914400}
}

@article{RODRIGUES2024100846,
title = {The social media Infodemic of health-related misinformation and technical solutions},
journal = {Health Policy and Technology},
volume = {13},
number = {2},
pages = {100846},
year = {2024},
issn = {2211-8837},
doi = {https://doi.org/10.1016/j.hlpt.2024.100846},
url = {https://www.sciencedirect.com/science/article/pii/S2211883724000091},
author = {Flinta Rodrigues and Richard Newell and Giridhara {Rathnaiah Babu} and Tulika Chatterjee and Nimrat Kaur Sandhu and Latika Gupta}
}

@article{vosoughi2018spread,
  author  = {Vosoughi, Soroush and Roy, Deb and Aral, Sinan},
  year    = {2018},
  title   = {The Spread of True and False News Online},
  journal = {Science},
  volume  = {359},
  number  = {6380},
  pages   = {1146--1151},
  date    = {2018-03-09},
  doi     = {10.1126/science.aap9559},
  publisher = {American Association for the Advancement of Science},
  issn    = {0036-8075},
  eissn   = {1095-9203}
}

@article{deFrancisco1992,
  author  = {De Francisco, Victoria Leto},
  year    = {1992},
  title   = {Deborah Tannen, You Just Don't Understand: Women and Men in Conversation. New York: William Morrow \& Co., 1990. Pp. 330},
  journal = {Language in Society},
  volume  = {21},
  number  = {2},
  pages   = {319--324},
  doi     = {10.1017/S0047404500015372}
}

@article{HOLMES1990185,
title = {Hedges and boosters in women's and men's speech},
journal = {Language \& Communication},
volume = {10},
number = {3},
pages = {185-205},
year = {1990},
issn = {0271-5309},
doi = {https://doi.org/10.1016/0271-5309(90)90002-S},
url = {https://www.sciencedirect.com/science/article/pii/027153099090002S},
author = {Janet Holmes}
}

@article{lakoff1973language,
  title={Language and woman's place},
  author={Lakoff, Robin},
  journal={Language in society},
  volume={2},
  number={1},
  pages={45--79},
  year={1973},
  publisher={Cambridge University Press}
}

@article{Daria2021CowDungCOVID,
  author  = {Daria, Sultana and Islam, Md. Rabiul},
  title   = {The use of cow dung and urine to cure COVID-19 in India: a public health concern},
  journal = {International Journal of Health Planning and Management},
  year    = {2021},
  doi     = {10.1002/hpm.3257}
}

@online{GulfNews2019CowUrine,
  author       = {{Gulf News}},
  title        = {Indian Health Minister Ashwini Kumar Choubey Is Working on Cow Urine to Prepare Medicines, India},
  year         = {2019},
  url          = {https://gulfnews.com/world/asia/india/indian-health-minister-ashwini-kumar-choubey-is-working-on-cow-urine-to-prepare-medicines-1.1567948447753},
  urldate      = {2021-06-12}
}

@article{Notermans2019Prayers,
  author  = {Notermans, Catrien},
  title   = {Prayers of Cow Dung: Women Sculpturing Fertile Environments in Rural Rajasthan (India)},
  journal = {Religions},
  year    = {2019},
  volume  = {10},
  number  = {2},
  pages   = {71},
  doi     = {10.3390/rel10020071},
  url     = {https://doi.org/10.3390/rel10020071}
}

@article{chen2024combating,
  title={Combating misinformation in the age of llms: Opportunities and challenges},
  author={Chen, Canyu and Shu, Kai},
  journal={AI magazine},
  volume={45},
  number={3},
  pages={354--368},
  year={2024},
  publisher={Wiley Online Library}
}

@article{tao2024cultural,
  title={Cultural bias and cultural alignment of large language models},
  author={Tao, Y. and Viberg, O. and Baker, R. S. and Kizilcec, R. F.},
  journal={PNAS Nexus},
  volume={3},
  number={9},
  pages={pgae346},
  year={2024},
  doi={10.1093/pnasnexus/pgae346}
}

@article{radford2022robust,
  title   = {Robust Speech Recognition via Large-Scale Weak Supervision},
  author  = {Radford, Alec and Kim, Jong Wook and Xu, Tao and Brockman, Greg and McLeavey, Christine and Sutskever, Ilya},
  journal = {arXiv preprint arXiv:2212.04356},
  year    = {2022},
  month   = dec,
  doi     = {10.48550/arXiv.2212.04356},
  url     = {https://arxiv.org/abs/2212.04356}
}

@article{Pawar2025CulturalAwarenessSurvey,
  title   = {Survey of Cultural Awareness in Language Models: Text and Beyond},
  author  = {Pawar, Siddhesh and Park, Junyeong and Jin, Jiho and Arora, Arnav and Myung, Junho and Yadav, Srishti and Haznitrama, Faiz Ghifari and Song, Inhwa and Oh, Alice and Augenstein, Isabelle},
  journal = {Computational Linguistics},
  year    = {2025},
  month   = sep,
  volume  = {51},
  number  = {3},
  pages   = {907--1004},
  doi     = {10.1162/COLI.a.14},
  url     = {https://direct.mit.edu/coli/article/51/3/907/130804/Survey-of-Cultural-Awareness-in-Language-Models}
}

@article{2025MisinformationPublicHealth,
  author    = {Denniss, Emily and Lindberg, Rebecca},
  title     = {Social media and the spread of misinformation: infectious and a threat to public health},
  journal   = {Health Promotion International},
  year      = {2025},
  volume    = {40},
  number    = {2},
  pages     = {daaf023},
  doi       = {10.1093/heapro/daaf023},
  publisher = {Oxford University Press}
}

@article{VanDerLinden2022Misinformation,
  title={Misinformation: susceptibility, spread, and interventions to immunize the public},
  author={Van Der Linden, Sander},
  journal={Nature medicine},
  volume={28},
  number={3},
  pages={460--467},
  year={2022},
  publisher={Nature Publishing Group US New York}
}

@article{VanDerLinden2025HealthMisinformation,
  title={Using psychological science to understand and fight health misinformation: An APA consensus statement.},
  author={Van der Linden, Sander and Albarrac{\'\i}n, Dolores and Fazio, Lisa and Freelon, Deen and Roozenbeek, Jon and Swire-Thompson, Briony and Van Bavel, Jay},
  journal={American Psychologist},
  year={2025},
  publisher={American Psychological Association}
}

@incollection{Alarabid2025LLMMisinformationDetection,
  author    = {Alarabid, A.},
  title     = {Leveraging Large Language Models for Misinformation Detection: A Focus on Public Health Misinformation on Social Media},
  booktitle = {Proceedings of the International Conference on Computing and Information Technology},
  year      = {2025},
  publisher = {Springer},
  doi       = {10.1007/978-981-96-9248-4_42}
}

@article{Peng2023PersuasiveHealthMisinformation,
  title={Persuasive strategies in online health misinformation: a systematic review},
  author={Peng, Wei and Lim, Sue and Meng, Jingbo},
  journal={Information, Communication \& Society},
  volume={26},
  number={11},
  pages={2131--2148},
  year={2023},
  publisher={Taylor \& Francis}
}

@article{Thakur2024YouTubeMisinformation,
  author  = {Thakur, Nirmalya and Cui, Shuqi and Knieling, Victoria and Khanna, Karam and Shao, Mingchen},
  title   = {Investigation of the Misinformation about COVID-19 on YouTube Using Topic Modeling, Sentiment Analysis, and Language Analysis},
  journal = {Computation},
  year    = {2024},
  volume  = {12},
  number  = {2},
  pages   = {28},
  doi     = {10.3390/computation12020028},
  publisher = {MDPI}
}

@article{Ma2025AIMisinformationLinguistic,
  title={Linguistic features of AI mis/disinformation and the detection limits of LLMs},
  author={Ma, Yulong and Zhang, Xinsheng and Ren, Jinge and Wang, Runzhou and Wang, Minghu and Chen, Yang},
  journal={Nature Communications},
  year={2025},
  publisher={Nature Publishing Group}
}

@article{Liu2025CulturalBiasLLM,
  author  = {Liu, Z.},
  title   = {Cultural Bias in Large Language Models: A Comprehensive Analysis and Mitigation Strategies},
  journal = {Journal of Transcultural Communication},
  year    = {2025},
  volume  = {3},
  number  = {2},
  pages   = {224--244},
  doi     = {10.1515/jtc-2023-0019},
  publisher = {De Gruyter}
}

@article{Chhikara2025PrismCulture,
  author  = {Chhikara, Garima and Kumar, Abhishek and Chakraborty, Abhijnan},
  title   = {Through the Prism of Culture: Evaluating LLMs' Understanding of Indian Subcultures and Traditions},
  journal = {arXiv preprint arXiv:2501.16748},
  year    = {2025},
  doi     = {10.48550/arXiv.2501.16748}
}

@inproceedings{Khandelwal2024IndianBhED,
  author    = {Khandelwal, Khyati and Tonneau, Manuel and Bean, Andrew M. and Kirk, Hannah Rose and Hale, Scott A.},
  title     = {Indian-BhED: A Dataset for Measuring India-Centric Biases in Large Language Models},
  booktitle = {Proceedings of the 2024 International Conference on Information Technology for Social Good},
  year      = {2024},
  pages     = {231--239},
  publisher = {Association for Computing Machinery},
  doi       = {10.1145/3677525.3678666}
}

@article{Essar2021CowDungCOVID,
  author  = {Essar, Mohammad Yasir and Kazmi, Syeda Kanza and Hasan, Mohammad Mehedi  and Costa, Ana Carla dos Santos and Ahmad, Shoaib},
  title   = {The rampant use of cow dung to treat COVID-19},
  journal = {Health Science Reports},
  year    = {2021},
  volume  = {4},
  number  = {4},
  pages   = {e363},
  doi     = {10.1002/hsr2.363},
  publisher = {Wiley}
}

@article{Hurford2022NarrativeMisinformationIndia,
  author  = {Hurford, Beth and Rana, Abhishek and Sachan, Rohan Samir Kumar},
  title   = {Narrative-based misinformation in India about protection against Covid-19: Not just another ``moo-point''},
  journal = {Indian Journal of Medical Ethics},
  year    = {2022},
  volume  = {7},
  number  = {1},
  pages   = {22--26},
  doi     = {10.20529/IJME.2021.050}
}

@article{Madathil2015YouTubeHealth,
  author    = {Madathil, Kapil Chalil and Rivera-Rodriguez, A. Joy and Greenstein, Joel S. and Gramopadhye, Anand K.},
  title     = {Healthcare information on YouTube: A systematic review},
  journal   = {Health Informatics Journal},
  year      = {2015},
  volume    = {21},
  number    = {3},
  pages     = {173--194},
  publisher = {SAGE Publications},
  doi       = {10.1177/1460458213512220}
}

@misc{Kemp2025DigitalGlobal,
  author       = {Kemp, Simon},
  title        = {Digital 2025: Global Overview Report},
  howpublished = {DataReportal},
  year         = {2025},
  url          = {https://datareportal.com/reports/digital-2025-global-overview-report}
}

@misc{Kemp2025DigitalIndia,
  author       = {Kemp, Simon},
  title        = {Digital 2025: India},
  howpublished = {DataReportal},
  year         = {2025},
  url          = {https://datareportal.com/reports/digital-2025-india}
}

@article{Aydin2025GenderDifferences,
  author  = {Aydın, Fatma},
  title   = {Examining Gender Differences in Social Media Language},
  journal = {Bulletin of Language and Literature Studies},
  volume  = {2},
  number  = {1},
  year    = {2025},
  month   = {Mar},
  doi     = {10.59652/blls.v2i1.519},
  url     = {https://journals.eikipub.com/index.php/blls/article/view/519}
}

@article{WangChen2022GenderSocialMedia,
  title={Gender performances on social media: A comparative study of three top key opinion leaders in China},
  author={Liu, Ming and Zhao, Ruinan and Feng, Jieyun},
  journal={Frontiers in psychology},
  volume={13},
  pages={1046887},
  year={2022},
  publisher={Frontiers Media SA}
}

@article{shang2025multitec,
  title={MultiTec: a data-driven multimodal short video detection framework for healthcare misinformation on TikTok},
  author={Shang, Lanyu and Zhang, Yang and Deng, Yawen and Wang, Dong},
  journal={IEEE Transactions on Big Data},
  volume={11},
  number={5},
  pages={2471--2488},
  year={2025},
  publisher={IEEE}
}

@inproceedings{perez2025detecting,
  title={Detecting Deception in Disinformation Across Languages: The Role of Linguistic Markers},
  author={P{\'e}rez-Montero, Alba and Gargova, Silvia and Lloret, Elena and Moreda, Paloma},
  booktitle={Proceedings of Recent Advances in Natural Language Processing (RANLP)},
  pages={943--952},
  year={2025}
}

@article{coli_veridicality,
  title={What if Deception cannot be Detected? A Cross-linguistic Study on the Limits of Deception Detection from Text},
  author={Velutharambath, Aswathy and Sassenberg, Kai and Klinger, Roman},
  journal={Computational Linguistics},
  pages={1--71},
  year={2026},
  publisher={MIT Press},
  doi={10.1162/COLI.a.614}
}

@inproceedings{danish2026comparative,
  title={A Comparative Study of mBERT and IndicBERT for Natural Language Processing in Indic Languages},
  author={Danish, M. and Liu, H. and Alshmrany, S.},
  booktitle={2025 IEEE 7th International Conference on Computing, Communication and Automation (ICCCA)},
  year={2025},
  publisher={IEEE}
}

@article{openai2023gpt4,
  title={GPT-4 Technical Report},
  author={OpenAI},
  year={2023},
  eprint={2303.08774},
  archivePrefix={arXiv},
  primaryClass={cs.CL},
  url={https://arxiv.org/abs/2303.08774}
}

@inbook{macfarland2016mann,
  author    = {MacFarland, Thomas W. and Yates, Jan M.},
  title     = {Mann–Whitney U Test},
  booktitle = {Introduction to Nonparametric Statistics for the Biological Sciences Using R},
  year      = {2016},
  publisher = {Springer International Publishing},
  address   = {Cham},
  pages     = {103--132},
  doi       = {10.1007/978-3-319-30634-6_4},
  isbn      = {978-3-319-30634-6}
}

@inproceedings{adilazuarda2024towards,
  title={Towards measuring and modeling “culture” in LLMs: A survey},
  author={Adilazuarda, Muhammad Farid and Mukherjee, Sagnik and Lavania, Pradhyumna and Singh, Siddhant Shivdutt and Aji, Alham Fikri and O’Neill, Jacki and Modi, Ashutosh and Choudhury, Monojit},
  booktitle={Proceedings of the 2024 Conference on Empirical Methods in Natural Language Processing},
  pages={15763--15784},
  year={2024}
}

\section{Appendix}

\begin{figure*}[t]
  \centering
  
   \includegraphics[width=0.8\linewidth]{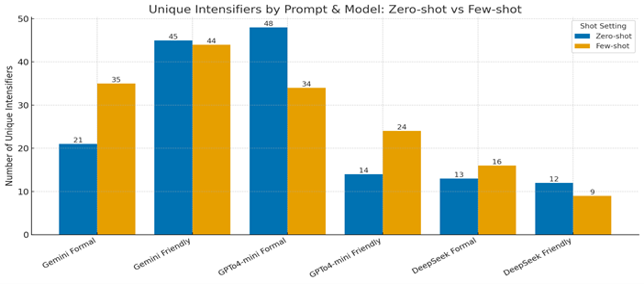}

   \caption{Comparison of the number of unique intensifiers generated across different models (Gemini, GPT4o-mini, and DeepSeek-V3.1) and prompt styles (formal vs. friendly), contrasting zero-shot and few-shot settings.}
   \label{fig:unique-identifiers}
\end{figure*}

\appendix

\label{sec:appendix_prompts}
\section{Prompts for RQ1}

\subsubsection{Scientific Terms Identification Prompt}

\newtcolorbox[auto counter]{PromptBox}{
  enhanced, breakable,
  width=\columnwidth,
  colback=gray!10, colframe=black,
  colbacktitle=black, coltitle=black,
  boxrule=0.5pt, arc=2mm,
  left=8pt, right=8pt, top=8pt, bottom=8pt,
  before skip=8pt, after skip=10pt
}

\newenvironment{PromptText}{%
  \ttfamily\footnotesize\linespread{1.03}\selectfont
  \setlength{\parindent}{0pt}%
  \raggedright 
  \setlength{\parskip}{4pt}
}{}

\newcommand{\Tag}[2]{\textbf{[#1]} #2}


\begin{PromptBox}
\begin{PromptText}

\textbf{User Instruction:}

You are an expert linguistic annotator specializing in scientific discourse. Carefully read the transcript below (ID: {transcript\_id}).

Extract ONLY terms that are explicitly scientific or pseudo-scientific—biomedical, chemical, laboratory, clinical, anatomical, or technical vocabulary typically associated with modern medicine or science.

\textbf{
Include:
}
\begin{itemize}
\item Specific chemicals, compounds, or medicines (e.g., enzyme, uric acid).
\item Diseases, medical conditions, or biological processes (e.g., cancer, immune response).
\item Diagnostic tools or procedures (e.g., MRI, ECG, blood test).
\item Scientific-sounding buzzwords used to lend credibility (e.g., antioxidants, toxins).

\end{itemize}

\textbf{
Exclude:
}
\begin{itemize}
\item  Ayurvedic or traditional medical terms.
\item Mythological or religious references.
\item General terms like 'health' or 'natural.

\end{itemize}

Return ONLY a comma-separated list of unique terms (no duplicates, no explanations). If none, return ''.

Transcript:
{transcript}

\end{PromptText}
\end{PromptBox}

\subsubsection{Traditional/Cultural Terms Identification Prompt}


\begin{PromptBox}
\begin{PromptText}

\textbf{User Instruction:}

You are an expert linguistic annotator specializing in traditional, Ayurvedic, and religious discourse. Carefully read the transcript below (ID: {transcript\_id}).

Extract ONLY traditional, Ayurvedic, mythological, religious, or culturally symbolic terms.

\textbf{
Include:
}
\begin{itemize}
\item Ayurvedic medicines, herbs, or therapies (e.g., Triphala, Ashwagandha).
\item Mythological or religious references (e.g., divine, Sanjeevani, sacred cow).
\item Traditional metaphors or culturally symbolic phrases (e.g., divine gift, Amrit).

\end{itemize}

\textbf{
Exclude:
}
\begin{itemize}
\item  Modern biomedical, chemical, or technical vocabulary.
\item General non-specific terms like "healthy" or "pure"

\end{itemize}

Return ONLY a comma-separated list of unique terms (no duplicates, no explanations). If none, return ''.

Transcript:
{transcript}

\end{PromptText}
\end{PromptBox}

\section{Prompts for RQ2}
\subsubsection{Formal Prompt}
\label{subsec:formal_prompt}






\begin{PromptBox}
\begin{PromptText}

\textbf{User Instruction:}

You are provided with a document containing transcripts from YouTube videos that discuss misinformation about \textit{gomutra} (cow urine). Carefully review the provided texts and complete the following tasks:

\textbf{Tasks:}

\begin{enumerate}
    \item Identify all intensifiers in each transcript.

\textbf{
Definition of an intensifier:
}
An intensifier is a word or a phrase specifically used to strengthen, exaggerate, or emphasize a claim. It is a lexical item that operates as a degree modifier on an adjective, an adverb, or occasionally a verb phrase. It augments the head it modifies, without altering the core semantic denotation of that head. For example, “very” (as in “The food was very good”), “strongly” (as in “I strongly believe in the power of education”), “too” (as in “The person was driving too fast“).

\item For each identified intensifier:
\begin{itemize}
\item Quote the exact intensifier.
\item  Identify the word it modifies.
\item Indicate its precise position within the text (e.g., beginning, middle, end of the sentence/paragraph).
\item Explain clearly how the intensifier amplifies or contributes to misinformation about \textit{gomutra}.

\end{itemize}

\item Summarize any patterns or common techniques observed across the transcripts regarding the use of intensifiers in promoting misinformation
\end{enumerate}

\end{PromptText}
\end{PromptBox}

\subsubsection{Friendly Prompt}
\label{subsec:friendly_prompt}





\begin{PromptBox}
\begin{PromptText}

\textbf{User Instruction:}

Hi Gemini 2.5 Pro! You're provided with a document containing transcripts from YouTube videos that discuss misinformation about \textit{gomutra} (cow urine). Please carefully review these texts and complete the following tasks to help us with qualitative analysis:

\textbf{Tasks:}

\begin{enumerate}
    \item Identify all intensifiers in each transcript.

\textbf{
Definition of an intensifier:
}
An intensifier is a word or a phrase specifically used to strengthen, exaggerate, or emphasize a claim. It is a lexical item that operates as a degree modifier on an adjective, an adverb, or occasionally a verb phrase. It augments the head it modifies, without altering the core semantic denotation of that head. For example, “very” (as in “The food was very good”), “strongly” (as in “I strongly believe in the power of education”), “too” (as in “The person was driving too fast“).

\item For each identified intensifier:
\begin{itemize}
\item Please quote the exact intensifier.
\item  Please identify the word it modifies.
\item Please indicate its precise position within the text (e.g., beginning, middle, end of the sentence/paragraph).
\item Please explain clearly how the intensifier amplifies or contributes to misinformation about \textit{gomutra}.

\end{itemize}

\item Summarize any patterns or common techniques you've noticed across the transcripts regarding the use of intensifiers to promote misinformation.
\end{enumerate}

Thanks for your help!

\end{PromptText}
\end{PromptBox}

\subsubsection{Few Shot Prompt}
\label{subsec:formal_prompt}






\begin{PromptBox}
\begin{PromptText}

The model was (a) first conditioned with a 5-transcript sample to steer its output before (b) processing all the transcripts.

\textbf{(a) User Instruction:}

Below are a few transcripts from YouTube videos that discuss misinformation about \textit{gomutra} (cow urine):

Transcript 1: [TEXT]

Transcript 2: [TEXT]

Transcript 3: [TEXT]

Transcript 4: [TEXT]

Transcript 5: [TEXT]

Carefully review the provided texts and complete the following tasks:

\textbf{Tasks:}

\begin{enumerate}
    \item Identify all intensifiers in each transcript.

\textbf{
Definition of an intensifier:
}
An intensifier is a word or a phrase specifically used to strengthen, exaggerate, or emphasize a claim. It is a lexical item that operates as a degree modifier on an adjective, an adverb, or occasionally a verb phrase. It augments the head it modifies, without altering the core semantic denotation of that head. For example, “very” (as in “The food was very good”), “strongly” (as in “I strongly believe in the power of education”), “too” (as in “The person was driving too fast“).

\item For each identified intensifier:
\begin{itemize}
\item Quote the exact intensifier.
\item  Identify the word it modifies.
\item Indicate its precise position within the text (e.g., beginning, middle, end of the sentence/paragraph).
\item Explain clearly how the intensifier amplifies or contributes to misinformation about \textit{gomutra}.

\end{itemize}

\item Summarize any patterns or common techniques observed across the transcripts regarding the use of intensifiers in promoting misinformation
\end{enumerate}

\textbf{(b) User Instruction:}

You are now provided with a document containing all the transcripts from YouTube videos that discuss misinformation about \textit{gomutra} (cow urine). Carefully review the provided texts and complete the following tasks:

\textbf{Tasks:}

\begin{enumerate}
    \item Identify all intensifiers in each transcript.

\textbf{
Definition of an intensifier:
}
An intensifier is a word or a phrase specifically used to strengthen, exaggerate, or emphasize a claim. It is a lexical item that operates as a degree modifier on an adjective, an adverb, or occasionally a verb phrase. It augments the head it modifies, without altering the core semantic denotation of that head. For example, “very” (as in “The food was very good”), “strongly” (as in “I strongly believe in the power of education”), “too” (as in “The person was driving too fast“).

\item For each identified intensifier:
\begin{itemize}
\item Quote the exact intensifier.
\item  Identify the word it modifies.
\item Indicate its precise position within the text (e.g., beginning, middle, end of the sentence/paragraph).
\item Explain clearly how the intensifier amplifies or contributes to misinformation about \textit{gomutra}.

\end{itemize}

\item Summarize any patterns or common techniques observed across the transcripts regarding the use of intensifiers in promoting misinformation
\end{enumerate}

\end{PromptText}
\end{PromptBox}

\subsection{Computational Environment and Compute Budget}
\label{subsec:compute_env}

\begin{itemize}
    \item \textbf{Compute Budget:} All experiments, data preprocessing, and API integrations were executed using a Google Colab Pro subscription (\textasciitilde\$10 USD/month).
    \item \textbf{Hardware Allocation:}Tasks were run on dynamically allocated high-RAM environments equipped with standard Pro-tier GPUs (predominantly NVIDIA V100 or T4). The project operated entirely within the standard allotted compute units, requiring no additional premium cloud instances or external clusters.
\end{itemize}

\subsection{Extended Transcription and Translation Protocols}
\label{subsec:dataset_protocols}

\begin{itemize}
    \item \textbf{Transcription Parameters:} We utilized Whisper's default decoding parameters (e.g., default temperature and beam search). To prevent language-switching artifacts, we explicitly forced the language parameter to Hindi (language="hi") and ran the model in FP32 precision (fp16=False).
    
    \item \textbf{Translation Parameters:} GPT-4 was run with a stochastic decoding temperature of 0.7. To mitigate this, the system prompt explicitly instructed the model to act as a culturally aware translator, strictly preserving the original meaning, tone, and rhetorical markers without paraphrasing. Furthermore, we preserve and release the original-language transcripts alongside the translations to enable future verification of these cross-lingual mappings.
   
\end{itemize}
 


\subsection{Evaluation Metrics Formulation}
\label{subsec:formulations}
Following evaluation metrics were calculated for each transcript via the following standard formulations:

\[ P = \frac{TP}{TP + FP} \]

\[ R = \frac{TP}{TP + FN} \]

\[ F1 = 2 \times \frac{P \times R}{P + R} \]

\begin{itemize}
    \item \textbf{True Positive ($TP$):} A scientific or traditional term correctly extracted by the LLM that perfectly aligns with the human annotation.
    \item \textbf{False Positive ($FP$):} A term incorrectly extracted by the LLM that was not present in the human-annotated ground truth.
    \item \textbf{False Negative ($FN$):} A valid term identified by the human annotator that the LLM failed to extract.
\end{itemize}







\end{document}